\documentclass{amia}
\usepackage[superscript,nomove]{cite}
\pdfoutput=1
\usepackage{lipsum} 
\usepackage{amsmath}
\usepackage{amsfonts}
\usepackage{amssymb}
\usepackage{subcaption} 

\usepackage{xcolor}
\usepackage{soul}
\usepackage[hidelinks]{hyperref}

\begin{document}

\title{Meta-Learning on Augmented Gene Expression Profiles for Enhanced Lung Cancer Detection}

\author{Arya Hadizadeh Moghaddam, MS$^{1}$, Mohsen Nayebi Kerdabadi, MS$^{1}$, \\Cuncong Zhong, PhD$^{1}$, Zijun Yao, PhD$^{1}$}

\institutes{
    $^1$University of Kansas, Lawrence, KS, USA
}

\maketitle

\section*{Abstract}

\textit{Gene expression profiles obtained through DNA microarray have proven successful in providing critical information for cancer detection classifiers. However, the limited number of samples in these datasets poses a challenge to employ complex methodologies such as deep neural networks for sophisticated analysis. To address this ``small data'' dilemma, Meta-Learning has been introduced as a solution to enhance the optimization of machine learning models by utilizing similar datasets, thereby facilitating a quicker adaptation to target datasets without the requirement of sufficient samples. In this study, we present a meta-learning-based approach for predicting lung cancer from gene expression profiles. We apply this framework to well-established deep learning methodologies and employ four distinct datasets for the meta-learning tasks, where one as the target dataset and the rest as source datasets. Our approach is evaluated against both traditional and deep learning methodologies, and the results show the superior performance of meta-learning on augmented source data compared to the baselines trained on single datasets. Moreover, we conduct the comparative analysis between meta-learning and transfer learning methodologies to highlight the efficiency of the proposed approach in addressing the challenges associated with limited sample sizes. Finally, we incorporate the explainability study to illustrate the distinctiveness of decisions made by meta-learning.}

\section*{Introduction}
Cancer is a leading cause of death in modern society.
For every 100,000 individuals in the United States, there were 47 new cases of lung cancer and 32 related deaths according to the statistics from CDC in 2020 \cite{cdc}.
DNA microarray, a method that measures the activity of tens of thousands of genes simultaneously, has emerged as an effective way of predicting cancer \cite{van2002gene}.
This technique enables the identification of the aberrant gene expression profile of cancer cells, offering valuable insights for early detection and personalized treatment of cancers of interest \cite{burczynski2000toxicogenomics,staunton2001chemosensitivity}.
Aiming for facilitating an efficient screening of a large set of gene expressions, DNA microarray is capable of analyzing over 48,000 transcript probes on a single chip \cite{ncbi}.
With such advantages, previous literature \cite{van2002gene,vadapalli2022artificial,pividori2023projecting} has extensively investigated using gene expression signatures to create data-driven classifiers to distinguish between cancer patients and controls.

Although gene expression profiling has become increasingly favored for predicting clinical outcomes, researchers and clinicians frequently face challenges in developing accurate predictive models due to the ``small data'' dilemma. This issue stems from the limited number of cancer patient samples typically available in clinical studies, often restricted by participant availability or budget constraints.
For instance, a study \cite{data255} enlisted 228 smokers to investigate the distinguishing signatures between 137 patients with Non-Small Cell Lung Cancer (NSCLC) and 91 patients with diagnosed benign nodules. In another study \cite{data830}, 95 subjects were recruited where 8 primary lung cancer patients were compared against 16 with tuberculosis, 25 with sarcoidosis, 8 with pneumonia, and the other control subjects.
Given the modest sample sizes usually ranging from several dozen to a few hundred participants, simpler predictive models, such as support vector machines that require fewer parameters, are often preferred for effective training. 
However, to comprehensively analyze the high-dimensional gene expression signatures and to further discover the intricate interplay among them, there is a growing demand for more sophisticated models like neural networks \cite{xie2016predictive}.
Therefore, how to apply deep learning methodologies with limited and high-dimensional gene expression data for predicting the outcome of interest becomes the research gap we attempt to address in this work.

In the exploration of predictive patterns from a small-scale gene expression dataset, one of the problems is whether we can harness the augmented data from additional gene expression studies to improve predictive accuracy. This question has been typically addressed through transfer learning, a two-stage learning pipeline, where models undergo pre-training on diverse datasets first to acquire a general-purpose learning foundation, and subsequently get fine-tuned on the target dataset to tailor their capabilities to a particular domain task. While transfer learning has proven beneficial in gene expression analysis \cite{lopez2020transfer}, its advantage still originates from leveraging abundant but unlabelled source datasets. Furthermore, transfer learning often demands considerable variability in downstream datasets to facilitate a seamless and non-overfitting adaptation to the target task. This requirement for extensive gene expression data during either the pre-training or fine-tuning phases still prevents transfer learning from meeting our goal: utilizing small-scale augmented gene expression data to enhance small-scale target data learning, to facilitate the application of deep learning frameworks.

To address this challenge, meta-learning, focusing on training the ability of ``learning to learn'', has emerged as an appropriate framework for differentiable models to facilitate the rapid adaptation to a target dataset using a small number of learning instances. Unlike transfer learning, meta-learning inherently optimizes for adaptability during the training phase\cite{qiu2020meta}, making it particularly effective for harnessing small-scale, and high-dimensional augmentation of gene expression to improve cancer detection.
In this work, we propose a Model-Agnostic Meta-Learning (MAML)\cite{finn2017model} based strategy of meta-learning to train sophisticated neural networks for the detection of lung cancer from gene expression samples. By training the capacity of adaptation using small-scale data, we aim to demonstrate the viability of deep learning applications in gene expression which often encounter limited sample availability.
Based on four real-world gene expression datasets related to lung cancer and other conditions, our findings validate the utility of integrating augmented data from disparate studies to refine model generalization.
With the help of augmented datasets, experimental results show that neural network models with meta-learning arrangement can consistently outperform either traditional or deep models developed de novo on a single dataset.

\section*{Data Description}
\begin{table}[t]
\small
\caption{Statistics of data.}
\label{table:statistical}
\centering
\begin{tabular}{ ccccc } 
\hline
Dataset  & \textbf{GSE13255} & \textbf{GSE135304} & \textbf{GSE12771} & \textbf{GSE42830}\\
\hline
\# of Samples            & 255      & 712       & 86       & 95       \\
\# of Features           & 15227    & 47323     & 48702    & 47323    \\
\# of Processed Features & 695      & 695       & 695      & 695      \\
Cancer Rate                  & 58.43\%  & 55.76\%   & 53.48\%  & 8.42\%  \\
\hline
\end{tabular}
\vspace{-0.3cm}
\end{table}

In our research, we use distinct gene expression datasets from four different studies. For each set of experiments, three of them serve as sources and the remaining is the target for meta-learning setting. Each dataset contains samples from gene expression levels obtained from microarrays with varying numbers of sensors. Below, the description of the datasets and their GEO ID are presented:

\textbf{GSE13255}\cite{data255}: This dataset contains samples from the University of Pennsylvania Medical Center between 2003 and 2007, consisting of subjects with a history of tobacco use and without previous lung cancer diagnosis, among whom some had benign non-calcified lung nodules and others were diagnosed with non-small-cell lung cancer (NSCLC). Blood samples were collected from all participants either during clinical visits or before surgery. This data is further extended with additional RNA samples from the New York University Medical Center.\\
\textbf{GSE135304}\cite{data304}:
The dataset contains samples from individuals found to have positive results on low-dose computed tomography scans at five clinical sites: the Helen F. Graham Cancer Center, The Hospital of the University of Pennsylvania, Roswell Park Comprehensive Cancer Center, Temple University Hospital, and participants from New York University Langone Medical Center, including those enrolled in an Early Detection Research Network lung screening program at New York University. The study includes individuals over 50 years old with a smoking history of over 20 pack-years and no cancer diagnosis within the past 5 years, except for non-melanoma skin cancer. Samples associated with malignant were collected within three months of diagnosis or prior to any invasive procedure.\\
\textbf{GSE12771}\cite{data771}: This dataset consists of blood samples obtained from smokers participating in epidemiological trials such as the European Prospective Investigation into Cancer and Nutrition, as well as the Cosmos and BC trials based in Cologne, Germany. All samples are extracted from patients either diagnosed with incident cases within the EPIC trial or prevalent cases within the Cosmos and BC trials. These samples contain a blood-based signature intended to differentiate between individuals diagnosed with lung cancer and unaffected smokers.\\
\textbf{GSE42830}\cite{data830}: The dataset in this study comprises patients diagnosed with various pulmonary diseases, including tuberculosis, sarcoidosis, pneumonia, and lung cancer, as well as healthy controls. The majority of TB patients were recruited from the Royal Free Hospital NHS Foundation Trust in London, while sarcoidosis patients were recruited from multiple hospitals in London, Oxford, and Paris between September 2009 and March 2012. Furthermore, Pneumonia patients were recruited from the Royal Free Hospital in London, and lung cancer patients and a subset of TB patients were recruited by the Lyon Collaborative Network in France. Exclusion criteria are significant medical history, age below 18, pregnancy, or immunosuppression.

Although the four different datasets are collected for different goals of the study, with different patient distributions, and having different class labels, we focus on binary classification for predicting whether a sample is associated with lung cancer. Moreover, we exclude samples from each dataset that lack completed expression arrays or are associated with an unidentified class. Statistical details about datasets are shown in Table \ref{table:statistical}.

\section*{Methodology}
\begin{figure}[t]
        \centering
	\includegraphics[width= 0.7\textwidth]{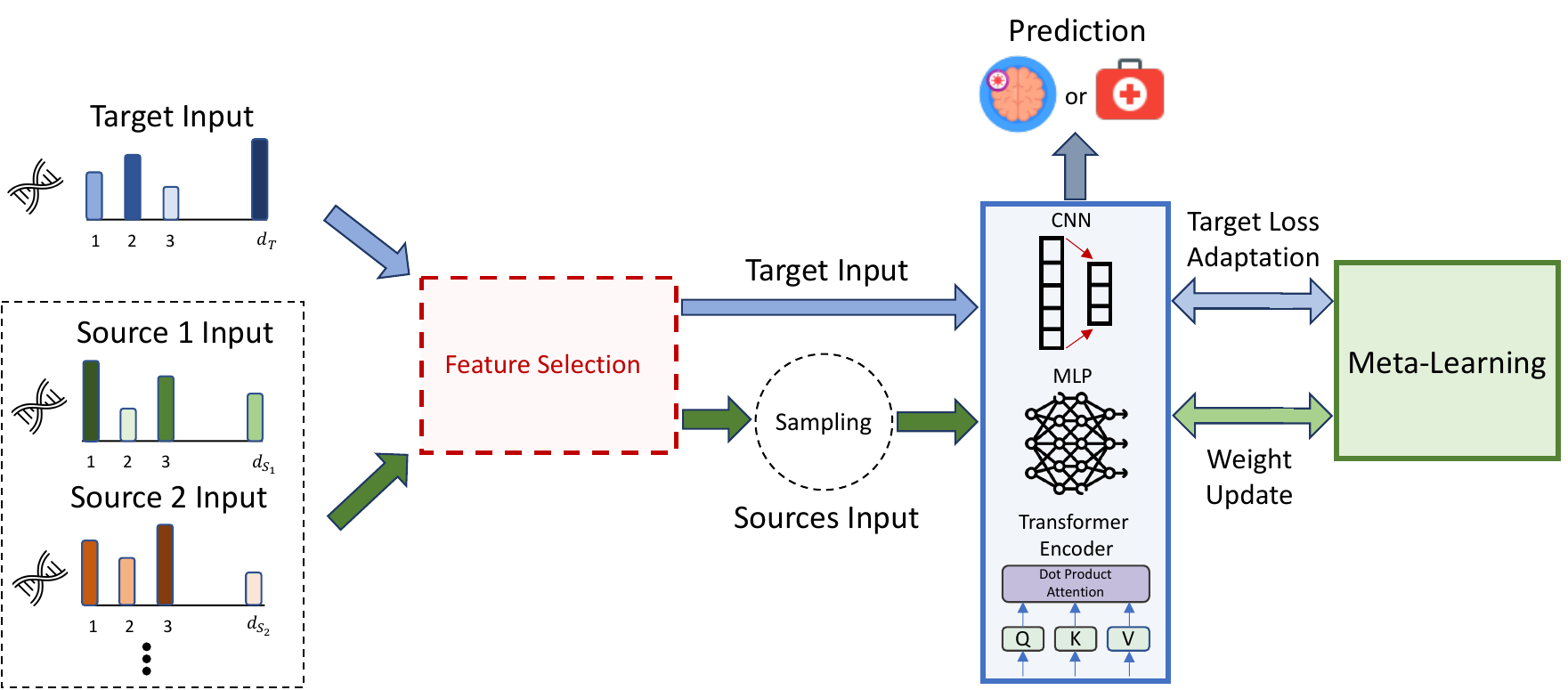}
	\caption{A visualization of the proposed method's architecture is presented. Initially, a feature selection process is employed to identify the most informative features while reducing dimensionality. Subsequently, the neural networks receive samples from source data and target batches, and loss values from the target and source datasets are used in meta-learning optimization.}\label{methodology}
\end{figure}

Meta-Learning \cite{metalearning} in the context of machine learning is known as ``learning-to-learn''. The concept involves using multiple datasets, referred to as ``source'' datasets, in the training process of the ``target'' dataset we aim to evaluate. Notably, the sources and target datasets have similarities with each other in their domain or prediction task. The meta-learning approach enables the model to update its weights based on diverse datasets and domains. Consequently, the neural networks gain enhanced generalization capabilities across both source and target datasets. The ultimate objective of meta-learning is to enrich the model designed for the target task with knowledge extracted from source datasets. This strategy was proved particularly advantageous when dealing with limited available data for the target task. By incorporating information from source datasets, the model achieves acceptable performance output with better results than just using the target dataset while disregarding the information embedded in the source datasets.

In this research, the datasets comprise gene expression levels obtained from Illumina sensors, with features indicating the gene expressions and labels denoting whether the sample exhibits cancer. Each dataset represents a distinct subset of individuals, such as smokers and non-smokers. To address this variability, we propose employing meta-learning as a mechanism for knowledge transfer between various distributions of samples. As illustrated in Figure \ref{methodology}, our methodology involves the initial collection of data from multiple source datasets, denoted as $\mathcal{D}_{\mathcal{S}i}$, where $i \in (1,2, ..., |\mathcal{S}|)$ and $\mathcal{|S|}$ represents the total number of source datasets. Additionally, a target dataset $\mathcal{D}_{\mathcal{T}}$ is incorporated into the analysis. Subsequently, a feature selection approach is employed for dimension adaptation, data normalization, and feature processing. Following feature selection, both source samples and target dataset are fed into a neural network for weight updates. To optimize this process, a meta-learning-based approach is proposed to enhance the model's generalizability across diverse tasks and help the model quickly adapt using a small number of data samples. Once the model is trained, it is utilized for predicting outcomes in the target task. In the following, we provide a detailed discussion about each sub-module of the proposed method.

\subsection*{Feature Selection}
Each sample in the datasets comprises a prediction label $y_\mathcal{M}$, showing if the cancer is detected or not, and a series of gene expression values $I_{j,\mathcal{M}} = (I^1_{j,\mathcal{M}}, I^2_{j,\mathcal{M}}, \cdots, I^{d_\mathcal{M}}_{j,\mathcal{M}})$, where $j$ is the index of the sample, $\mathcal{M}$ represents either the source data, $D_{\mathcal{S}i}$, or the target dataset, $D_{\mathcal{T}}$, with $d_\mathcal{M}$ being the number of features for dataset $\mathcal{M}$. The variability in $d_\mathcal{M}$ across datasets poses a challenge for the neural network architecture during training. Hence, utilizing a feature selection approach becomes crucial to identify features that (1) are shared among all datasets and (2) contain essential information for capturing cancer patterns by the classifier.

To address these objectives, we first identify the common genes detected across all datasets, next, we leverage the BioGrid \cite{biogrid} database to identify the genes that interact with other genes. Empirically, we found that focusing on the genes involved in genetic interactions increases the likelihood of capturing meaningful embedding. The feature selection process is formalized in Equation \ref{biogird}, where $I^{bg}_{\mathcal{M}} \in \mathbb{R}^{d_\text{fs}}$, and $d_\textit{fs}$ represents the reduced dimension.
\begin{equation}
\label{biogird}
I^{bg}_{\mathcal{M}}= \text{BioGrid}(I_{\mathcal{M}})
\end{equation}
Next, we employ standard normalization to obtain normalized features to enhance embedding learning as per Equation \ref{eq:normalized}, where $\mu_{\mathcal{M}}$ represents the mean of $I^{bg}_{\mathcal{M}}$ samples, and $\sigma^{bg}_{\mathcal{M}}$ is the standard deviation.
\begin{equation}
\label{eq:normalized}
\hat{I}_{j,\mathcal{M}} = \frac{{I^{bg}_{j,\mathcal{M}} - \mu^{bg}_{\mathcal{M}}}}{\sigma^{bg}_{\mathcal{M}}}
\end{equation}
Therefore, by applying Equations \ref{biogird} and \ref{eq:normalized}, we extract datasets $\hat{\mathcal{D}}_{\mathcal{T}}$ and $\hat{D}_{\mathcal{S}_i}$ for $i \in (1, 2, \cdots, |S|)$ from $\hat{I}_{\mathcal{M}}$. These processed datasets are subsequently employed in the deep learning model for further analysis.

\begin{figure}[t]
        \centering
	\includegraphics[width= 0.6\textwidth]{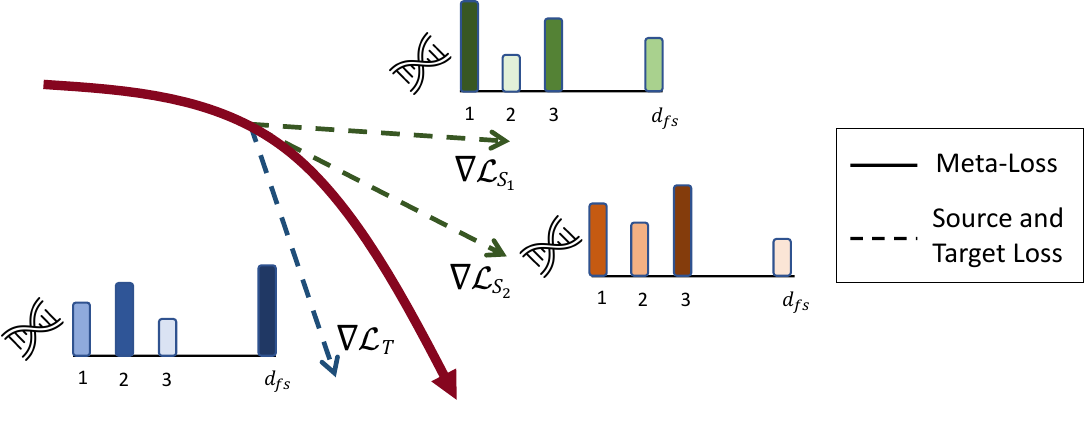}
 \vspace{-10pt}
	\caption{Illustration of the meta-loss adaption with the sources and target losses. A superior convergence of the loss function across diverse datasets with improve generalization is achieved through meta-learning.}\label{metls}
\end{figure}%

\subsection*{Classifier}
After obtaining processed data containing selected features for all datasets, the next step involves employing a classifier to extract embeddings from these features layer by layer. These embeddings are then utilized to predict whether the input sample corresponds to cancer or not. In this study, we use Multi-Layer Perceptron (MLP), Convolutional Neural Networks (CNNs), and Transformer's Encoder models for the meta-learning task. Below, we provide the formulations for all approaches:

\textbf{Multi-Layer Perceptron (MLP):} The MLP method comprises multiple linear layers to output a prediction score. Despite its apparent simplicity, MLPs prove effective when handling datasets with a small number of samples. The formulation for a linear layer is demonstrated in Equation \ref{eq:MLP}, where $b$ denotes the bias term, LeakyRelu is the activation function, $W_{\text{MLP}} \in \mathbb{R}^{d_{\text{output}} \times d_{\text{input}}}$, $d_{\text{output}}$ represents the output dimension of the layer, and $d_{\text{input}}$ represents the input dimension.
\begin{equation}
\label{eq:MLP}
h = \text{LeakyReLU}(W_{\text{MLP}} x + b)
\end{equation}
The stacked linear layers are followed by an output layer with a sigmoid activation applied to obtain the prediction value, as shown in Equation \ref{eq:cnn_linear}, where $W_\text{output} \in \mathbb{R}^{1 \times d_\text{last}}$, $d_\text{last}$ equals to the output dimension of the last hidden layer, $h_\text{final} \in$ $\mathbb{R}^{d_\text{last}}$ and $\hat{y}$ represents the prediction score between 0 and 1. A score closer to 1 indicates a higher likelihood of the sample having cancer.
\begin{equation}
\label{eq:cnn_linear}
\hat{y}=\text{Sigmoid}(W_\text{output} h_\text{final})
\end{equation}
\textbf{Convolutional Neural Networks (CNNs):} CNNs have been applicable in extracting meaningful features through the multiple layers of filters with kernels for prediction tasks. 
Typically, CNN architectures incorporate convolutional layers followed by pooling layers to downsample, reducing dimensionality and overfitting risks. The mathematical formulation for convolutional layer is shown in the Equation \ref{eq:cnn}, where $p$ denotes windows size, $W_{c, k} \in \mathbb{R}^{c_\text{in} \times c_\text{out} \times p}$, $c_\text{in}$ and  $c_\text{out}$ is the number of input and output channels, $h_j$ represents the output representation at position $j$, and $x$ denotes the input of the layer.
\begin{equation}
\label{eq:cnn}
h_j=\sum_{c=0}^{c_\text{in}-1} \sum_{k=1}^p W_{c, k} x
_{c, j-k} 
\end{equation}
Subsequently, the convolutional layer is followed by a max-pooling layer, represented by Equation \ref{eq:maxpool}, where $g$ denotes the pool size.
\begin{equation}
\label{eq:maxpool}
 h_\text{pooled}=\max \left\{h[i: i+g]\right\}
\end{equation}

The convolutional layers followed by max-pooling layers are stacked together for multi-level feature extraction. Finally, Similar to Equation \ref{eq:cnn_linear}, a linear layer is employed for the output prediction.

\textbf{Transformer's Encoder:} Self-attention in the Transformer architecture \cite{attentionall}, has shown efficiency in finding interdependencies among input features and has been used for healthcare prediction tasks \cite{nayebi2023contrastive}. In our dataset, where gene expression sensors may exhibit correlations, this mechanism can yield more informative representations. The process begins by extracting three vectors: Query, Key, and Value, as shown in Equation \ref{eq:qkv} where $x$ represents the 1D input of the Transformer, while $Q$, $K$, and $V$ belong to $\mathbb{R}^d$, with $d$ equals to the embedding dimension.
\begin{equation}
\label{eq:qkv}
Q = {W}_\text{query}x, \quad K = {W}_\text{key} x, \quad V = {W}_\text{value} x
\end{equation} 
Then, an attention matrix is developed showing the co-dependency between the input features, and finally by using $V$ vector the output is attained by using the Equation \ref{eq:transformer_output}, where $h_{\text{attention}}$ $\in \mathbb{R}^d$ is the output representation of the Encoder.
\begin{equation}
\label{eq:transformer_output}
h_{\text{attention}}=\operatorname{\text{Softmax}}(\frac{ Q K^{T}}{\sqrt{d}}) V
\end{equation}
Finally, the encoder's prediction score is fed into a linear layer similar to Equation \ref{eq:cnn_linear}.

\subsection*{Loss Function} The mentioned neural networks have various parameters $\Theta$, which undergoes updates throughout the training process. $\hat{I}_{\mathcal{M}}$ is the input of the networks and the output is determined using Equation \ref{eq:output}, where $\hat{\mathbf{y}}$ represents the predicted score related to lung cancer.
\begin{equation}
\label{eq:output}
\hat{\mathbf{y}}_j=f\left(\hat{I}_{j,\mathcal{M}} ; \Theta\right)
\end{equation}
The predicted output is further utilized in a binary cross entropy loss using Equation \ref{eq:bce}, where $N_{\mathbf{M}}$ denotes the total number of samples in a batch and $\mathcal{L}(\Theta)$ equals to the loss value.
\begin{equation}
\label{eq:bce}
\mathcal{L}(\Theta)=-\frac{1}{N_{\mathcal{M}}} \sum_{i=1}^N\left(\left(\mathbf{y}_j\right) \log \left(\hat{\mathbf{y}}_j\right)+\left(1-\mathrm{y}_j\right) \log \left(1-\hat{\mathbf{y}}_j\right)\right)
\end{equation}

\subsection*{Meta-Learning}
Inspired by MedPred \cite{zhang2019metapred}, MetaCare++ \cite{tan2022metacare++} and Model-Agnostic Meta-Learning (MAML) approaches \cite{finn2017model}, this research proposes a meta-learning approach to enhance the optimization of neural network weights with both source datasets and the target dataset. The proposed approach contributes to the development of a generalized model, especially when the target dataset has a limited number of samples.

The process initiates with a uniform random sampling of a batch from all source datasets. Next, for each source batch, new weight values are extracted using Equation \ref{eq:sourceupdate}, where $\alpha$ is the learning rate, and $\Theta^*_{\mathcal{S}i}$ denotes the updated weights for source $j$. Stochastic Gradient Descent (SGD) is utilized for the optimization due to the statistical structure of the process.
\begin{equation}
\label{eq:sourceupdate}
\Theta^*_{\mathcal{S}i}=\Theta- \alpha \nabla \mathcal{L}_\Theta\left((\hat{I}_{\mathcal{S}i}, \mathrm{y}_{\mathcal{S}i}) \in \hat{\mathcal{D}}_{\mathcal{S}i}\right)
\end{equation}
The updated set of weights for each source dataset is used to compute the overall source loss value, as defined in Equation \ref{equation:losses}, where $\mathcal{L}^{\mathcal{S}}_{\Theta^*}$ contains the influence of sources datasets, and $\mathrm{y}_{\mathcal{S}i}$ denotes the cancer label.

\begin{equation}
\label{equation:losses}
\mathcal{L}^{\mathcal{S}}_{\Theta^*}=\frac{1}{|\mathcal{S}|} \sum_{\left(\mathrm{X}_i, \mathrm{y}_i\right) \in \hat{\mathcal{D}}_{\mathcal{S}j}} \mathcal{L}_{\Theta^*_{\mathcal{S}i}}\left((\hat{I}_{\mathcal{S}i}, \mathrm{y}_{\mathcal{S}i}) \in \hat{\mathcal{D}}_{\mathcal{S}i}\right)
\end{equation}
While $\mathcal{L}^{\mathcal{S}}_{\Theta^*}$ encapsulates information from the entire source dataset, it lacks consideration for the adaptability of the target dataset. To address this, we initially compute the loss value specific to the target dataset batch, employing Equation \ref{eq:targetloss}. Here, $\mathcal{L}_{_\Theta}^{\mathcal{T}}$ represents the loss value corresponding to the target dataset. Notably, the model without any meta-learning would consider using only this loss value for the optimization. 
\begin{equation}
\label{eq:targetloss}
\mathcal{L}^{\mathcal{T}}_\Theta=\mathcal{L}_{\Theta}\left((\hat{I}_{\mathcal{T}}, \mathrm{y}_{\mathcal{T}}) \in \hat{\mathcal{D}}_{\mathcal{T}}\right)
\end{equation}
To combine the influences of both source and target datasets, the ultimate meta-loss is computed using Equation \ref{eq:metaloss}, with $\lambda$ representing a scaling factor to regulate the impact of the source loss, and $\mathcal{L}^\textit{Meta}$ denotes the weighted final loss value.
\begin{equation}
\label{eq:metaloss}
\mathcal{L}^\textit{Meta} = \lambda \mathcal{L}^{\mathcal{T}}_\Theta + (1-\lambda) \mathcal{L}^{\mathcal{S}}_{\Theta^*}
\end{equation}
Finally, the meta-loss is employed to optimize the weights of the proposed method using Equation \ref{eq:finalopt}, where $\beta$ is the learning rate, and the Adam optimizer is utilized for the optimization. This final optimization process refines the model parameters based on the comprehensive meta-loss value, enhancing the overall generalization of the proposed approach. For clarity, we present the evolution of the loss values in Figure \ref{metls}. The figure illustrates that the change in loss values is a combination of various source losses and the target loss. 
\begin{equation}
\label{eq:finalopt}
\Theta=\Theta-\beta \nabla \mathcal{L}^\textit{Meta}
\end{equation}
Our proposed approach offers two primary advantages: (1) in scenarios with limited data, the source loss aids the model in converging towards a more generalized network by decreasing the impact of noise and addressing the low-resource dilemma; (2) within the scope of this research, each dataset has samples from different types of patients and the meta-learning approach proves valuable in guiding the model toward achieving broad applicability across all patients with various health conditions.

\begin{table}[t]
\caption{Performance evaluation of lung cancer prediction task using GSE13255 as target and others as sources.}
\label{table:255}
\small
\centering
\begin{tabular}{cccccc}
\hline
\textbf{Model}      & \textbf{Accuracy} & \textbf{F1 score} & \textbf{Precision} & \textbf{Recall} & \textbf{PRAUC}  \\
\hline
Logistic Regression & 0.7409            & 0.7207            & 0.7302             & 0.7280           & 0.8222          \\
Decision Tree       & 0.6354            & 0.6004            & 0.6378             & 0.6217          & 0.7022          \\
SVM                 & 0.7682            & 0.7459            & 0.7624             & 0.7495          & 0.8651          \\
Random Forest       & 0.7809            & 0.7633            & 0.7991             & 0.7764          & 0.8551          \\
\hline
MLP                 & 0.8388            & 0.8285           & 0.8385   & 0.8330    & 0.8937        \\
CNN                 & 0.8314     & 0.8177      & 0.8328   & 0.8205     & 0.8817       \\
Transformer       & 0.8438   & 0.8345  & 0.8405  & 0.8360 & 0.9084  \\

\hline
\textbf{MLP + Meta-Learning} & 0.8626 & 0.8538  & 0.8609   & 0.8594  & 0.9283  \\
\textbf{CNN + Meta-Learning} & 0.8629 & 0.8553   & 0.8639        & 0.8592  & 0.9142 \\
\textbf{Transformer + Meta-Learning} & \textbf{0.8785}   & \textbf{0.8730}  & \textbf{0.8826}    &  \textbf{0.8822} & \textbf{0.9298}

\\    
\hline
\end{tabular}
\end{table}

\begin{table}[t]
\vspace{0.3cm}
\caption{Performance evaluation of lung cancer prediction task using GSE135304 as target and others as sources.}
\label{table:304}
\small
\centering
\begin{tabular}{cccccc}
\hline

\textbf{Model}      & \textbf{Accuracy} & \textbf{F1 score} & \textbf{Precision} & \textbf{Recall} & \textbf{PRAUC}  \\
\hline
Logistic Regression & 0.6841            & 0.6747            & 0.6793             & 0.6758          & 0.7784          \\
Decision Tree       & 0.6223            & 0.6115            & 0.6224             & 0.6205          & 0.6668          \\
SVM                 & 0.6882            & 0.6785            & 0.6879             & 0.6809          & 0.7804          \\
Random Forest       & 0.6799            & 0.6643            & 0.6829             & 0.6682          & 0.7587          \\
\hline
MLP                 & 0.7360              & 0.7256     & 0.7365  & 0.7288    & 0.8082       \\
CNN                 & 0.7290            & 0.7112          & 0.7315 & 0.7261    & 0.7794       \\
Transformer                 & 0.7263   & 0.7203  & 0.7262    & 0.7246 & 0.7919  \\
\hline

\textbf{MLP + Meta-Learning}  & 0.7585   & 0.7505  & 0.7530  & 0.7511 & 0.8220 \\
\textbf{CNN + Meta-Learning} & 0.7402            & 0.7304       & 0.7386    & 0.7293    & 0.7838 \\    
\textbf{Transformer + Meta-Learning} &  \textbf{0.7668}    & \textbf{0.7580}    & \textbf{0.7686}    & \textbf{0.7574}  & \textbf{0.8239} 
\\
\hline        
\end{tabular}
\end{table}

\begin{table}[t]
\vspace{0.2cm}
\caption{Performance evaluation of lung cancer prediction task using GSE12771 as target and others as sources.}
\label{table:771}
\centering
\small
\begin{tabular}{cccccc}
\hline
\textbf{Model}      & \textbf{Accuracy} & \textbf{F1 score} & \textbf{Precision} & \textbf{Recall} & \textbf{PRAUC}  \\
\hline
Logistic Regression & 0.6264            & 0.5836            & 0.6025             & 0.6238          & 0.6572          \\
Decision Tree       & 0.6417            & 0.5946            & 0.6182             & 0.6430           & 0.6240           \\
SVM                 & 0.6389            & 0.6050             & 0.6448             & 0.6487          & 0.7461          \\
Random Forest       & 0.7472            & 0.7245            & 0.7262             & 0.7551          & 0.8584          \\
\hline
MLP                 & 0.8264            & 0.8104            & 0.8390              & 0.8242          & 0.8436          \\
CNN                 & 0.7431            & 0.7327            & 0.7733             & 0.7871          & 0.8291          \\
Transformer                 & 0.7958  & 0.7543 & 0.7604 & 0.7579  & 0.8555  \\

\hline
\textbf{MLP + Meta-Learning} & 0.8613   & 0.8552            & 0.8737    & 0.8650          & \textbf{0.9118} \\
\textbf{CNN + Meta-Learning} & 0.8611              & 0.8546   & 0.8633             & 0.8719 & 0.8558  \\  
\textbf{Transformer + Meta-Learning} & \textbf{0.8750}   & \textbf{0.8608}  & \textbf{0.8957}    & \textbf{0.8717}   & 0.8626
\\
\hline        
\end{tabular}
\vspace{-0.3cm}
\end{table}

\begin{table}[t]
\vspace{10pt}
\caption{Performance evaluation of lung cancer prediction task using GSE42830 as target and others as sources.}
\label{table:830}
\small
\centering
\begin{tabular}{cccccc}
\hline
\textbf{Model}      & \textbf{Accuracy} & \textbf{F1 score} & \textbf{Precision} & \textbf{Recall} & \textbf{PRAUC}  \\
\hline
Logistic Regression & 0.9167            & 0.6774            & 0.6583             & 0.7000             & 0.4143          \\
Decision Tree       & 0.8222            & 0.5684            & 0.5755             & 0.5681          & 0.0900            \\
SVM                 & 0.9189            & 0.7274            & 0.7094             & 0.7500             & 0.4513          \\
Random Forest       & 0.8922          & 0.6204            & 0.6072             & 0.6389             & 0.2847          \\
\hline
MLP                 & 0.9456            & 0.8182            & 0.8276             & 0.8194          & 0.3268            \\
CNN                 & 0.9144            & 0.6767            & 0.6572             & 0.7000             & 0.1311          \\
Transformer                 & 0.9378 & 0.7576 & 0.7678 & 0.7667 & 0.3378 \\

\hline
\textbf{MLP + Meta-Learning} & 0.9478   & 0.8599   & 0.8526    & 0.8688  & 0.3804 \\
\textbf{CNN + Meta-Learning} & 0.9167            & 0.7764            & 0.7583             & 0.8000             & 0.2123 \\    
\textbf{Transformer + Meta-Learning} & \textbf{0.9689}   & \textbf{0.9244} & \textbf{0.9338}   & \textbf{0.9250}    & \textbf{0.5533}   \\
\hline
\end{tabular}
\end{table}

\section*{Evaluation}

To show the effectiveness of meta-learning for gene expression mining, we aim to answer the following questions:

\begin{itemize}
\vspace{-10pt}
\item How does the performance of the proposed meta-learning approach compare with traditional statistical machine-learning methodologies across the datasets?
\item How does the integration of meta-learning contribute to the performance of deep learning models that have not been trained using meta-learning techniques?
\item What are the comparative advantages of meta-learning over transfer learning strategy in terms of classification performance?
\vspace{-10pt}
\end{itemize}

\begin{table}[t]
\vspace{0.2cm}
\caption{Comparison between meta-learning and transfer learning on GSE13255 as target and others as sources.}
\label{table:t255}
\small
\centering
\begin{tabular}{cccccc}
\hline
\textbf{Model}      & \textbf{Accuracy} & \textbf{F1 score} & \textbf{Precision} & \textbf{Recall} & \textbf{PRAUC}  \\
\hline
MLP + Transfer Learning                 & 0.8389            & 0.8280            & 0.8424             & 0.8254           & 0.8998          \\
CNN + Transfer Learning                 & 0.8358            & 0.8187            & 0.8399             & 0.8217          & 0.8807         \\
Transformer + Transfer Learning                 & 0.8554   & 0.8476 & 0.8573    & 0.8503  & 0.8991 \\

\hline
\textbf{MLP + Meta-Learning} & 0.8626   & 0.8538    & 0.8609             & 0.8594 & 0.9283           \\
\textbf{CNN + Meta-Learning} & 0.8629   & 0.8553    & 0.8639             & 0.8592 & 0.9142\\
\textbf{Transformer + Meta-Learning} & \textbf{0.8785}   & \textbf{0.8730}  & \textbf{0.8826}    &  \textbf{0.8822} & \textbf{0.9298}

\\    
\hline
\end{tabular}
\vspace{-0.4cm}
\end{table}

\begin{table}[t]
\vspace{10pt}
\caption{Performance comparison between meta-learning and transfer learning using GSE135304 as target and others as sources.}
\label{table:t304}
\small
\centering
\begin{tabular}{cccccc}
\hline

\textbf{Model}      & \textbf{Accuracy} & \textbf{F1 score} & \textbf{Precision} & \textbf{Recall} & \textbf{PRAUC}  \\
\hline

MLP + Transfer Learning                 & 0.7431            & 0.7394            & 0.7427             & 0.7438           & 0.7933          \\
CNN + Transfer Learning                 & 0.7163            & 0.7116            & 0.7274             & 0.7213           & 0.7773          \\
Transformer + Transfer Learning                 & 0.7402   & 0.7340 & 0.7347    & 0.7362  & 0.7911 \\
\hline
\textbf{MLP + Meta-Learning}  & 0.7585   & 0.7505    & 0.7530    & 0.7511 & 0.8220 \\
\textbf{CNN + Meta-Learning} & 0.7402            & 0.7304            & 0.7386             & 0.7293          & 0.7838 \\    
\textbf{Transformer + Meta-Learning} &  \textbf{0.7668}   & \textbf{0.7580}   & \textbf{0.7686}    & \textbf{0.7574} & \textbf{0.8239} 
\\
\hline        
\end{tabular}
\end{table}
\subsection*{Experiment Design}
To address the following questions, we conducted experiments to validate the performance of our proposed method. In each experiment, one dataset is used as a target, and others are utilized as source datasets in the context of meta-learning. Our evaluation methodology employed 10-fold cross-validation, wherein 10\% of the samples were reserved for testing and the remaining 90\% were utilized for training. In each metric, we calculate the average of the values over all of the folds. For the CNN architecture, we employed two layers of convolution and max-pooling. The convolutional layers use a kernel size of 3, a stride of 1, and padding set to 1. The max-pooling layers have a kernel size of 2 and a stride of 2. In MLP, we employed two hidden linear layers, and in Transformer's encoder, we utilized one layer of self-attention mechanism. All methods are trained using $\alpha$ equal to 0.0004 for meta-learning, with a momentum of 0.2, and $\beta$ equals 0.0004 for total optimization. The models are trained for 40 epochs. Additionally, we explored a greedy approach to determine optimal hyperparameters, including number of convolutional channels (chosen from values 32, 64, and 128), batch size (selected from values 32 and 64), embedding dimension (chosen from values 32, 64, and 128), and $\lambda$ values (selected from 0.1, 0.3, 0.5, 0.7, 0.9, and 1). The best-performing values for each dataset are reported. In evaluating our model's performance, we employ five essential classification metrics: Accuracy, Precision, Recall, F1-score, and PRAUC (area under the precision-recall curve). We release the implementation code\footnote{\url{https://github.com/aryahm1375/MetaGene}.} on Github.


\subsection*{Baselines} We conduct performance comparisons using both statistical and deep learning baseline methods. Our statistical baseline methods include the following:\\
\textbf{Logistic Regression} is a statistical model to predict the probability of an event occurring based on input features, and employing a logistic function to map inputs to the probability.\\
\textbf{Decision Tree} is a hierarchical model for decision-making, where each internal node represents a decision based on an input feature, leading to classifications represented by leaf nodes.\\
\textbf{Support Vector Machine (SVM)} classifies data by finding the hyperplane that best separates different classes in the dataset while maximizing the margin between them.\\
\textbf{Random Forest} is an ensemble learning method that constructs multiple decision trees and combines their prediction score to improve accuracy and robustness in classification and regression tasks.\\
In addition to the statistical method, we employ \textbf{MLP}, \textbf{CNN}, and \textbf{Transformer's encoders} as baseline models for the prediction task without using the meta-learning technique.

\begin{figure}[t]
\centering
\begin{subfigure}[t]{.24\textwidth}
  \centering
  \includegraphics[width=\linewidth]{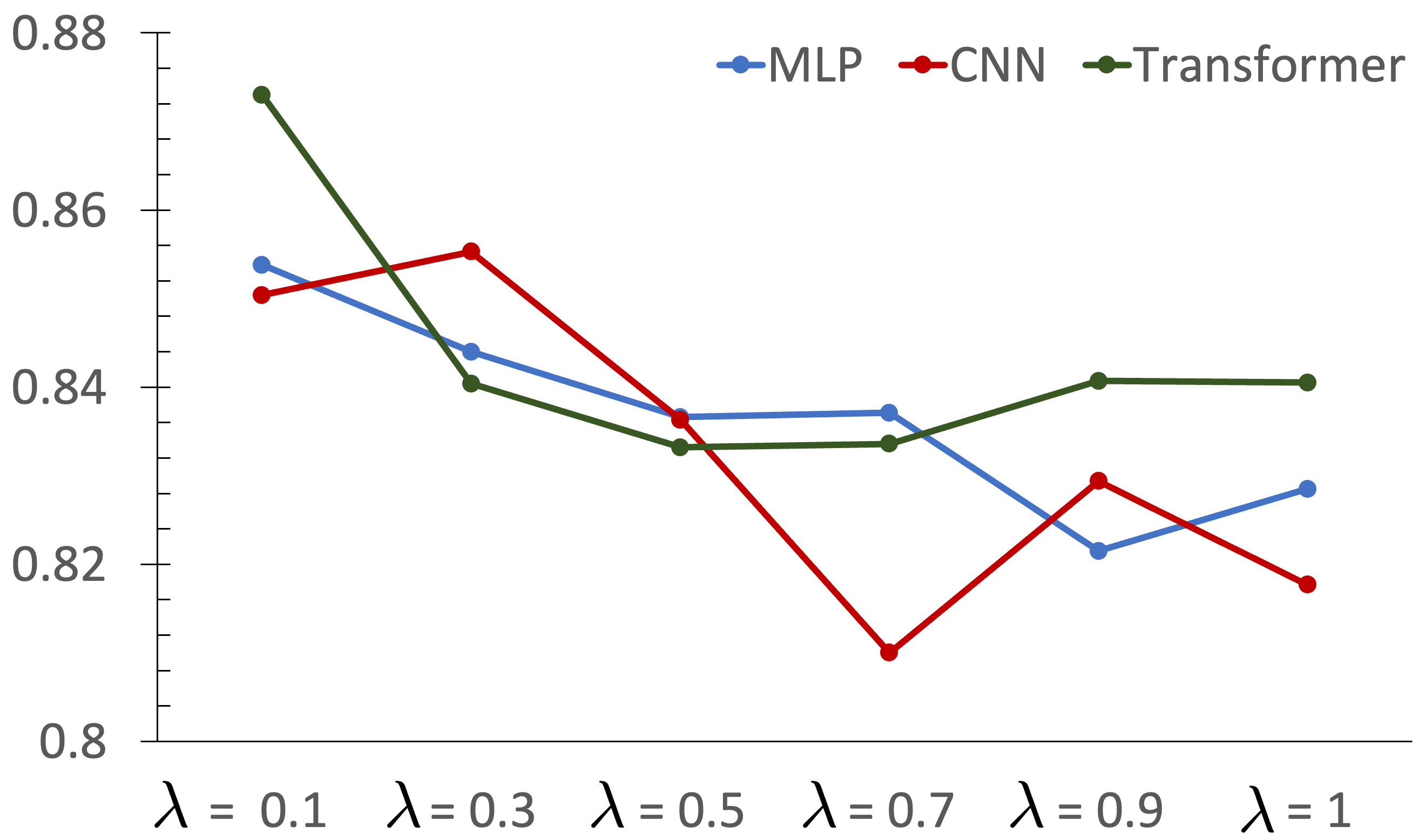}  
  \caption{GSE13255}
  \label{fig:ds_255}
\end{subfigure}
\begin{subfigure}[t]{.24\textwidth}
  \centering
  \includegraphics[width=1\linewidth]{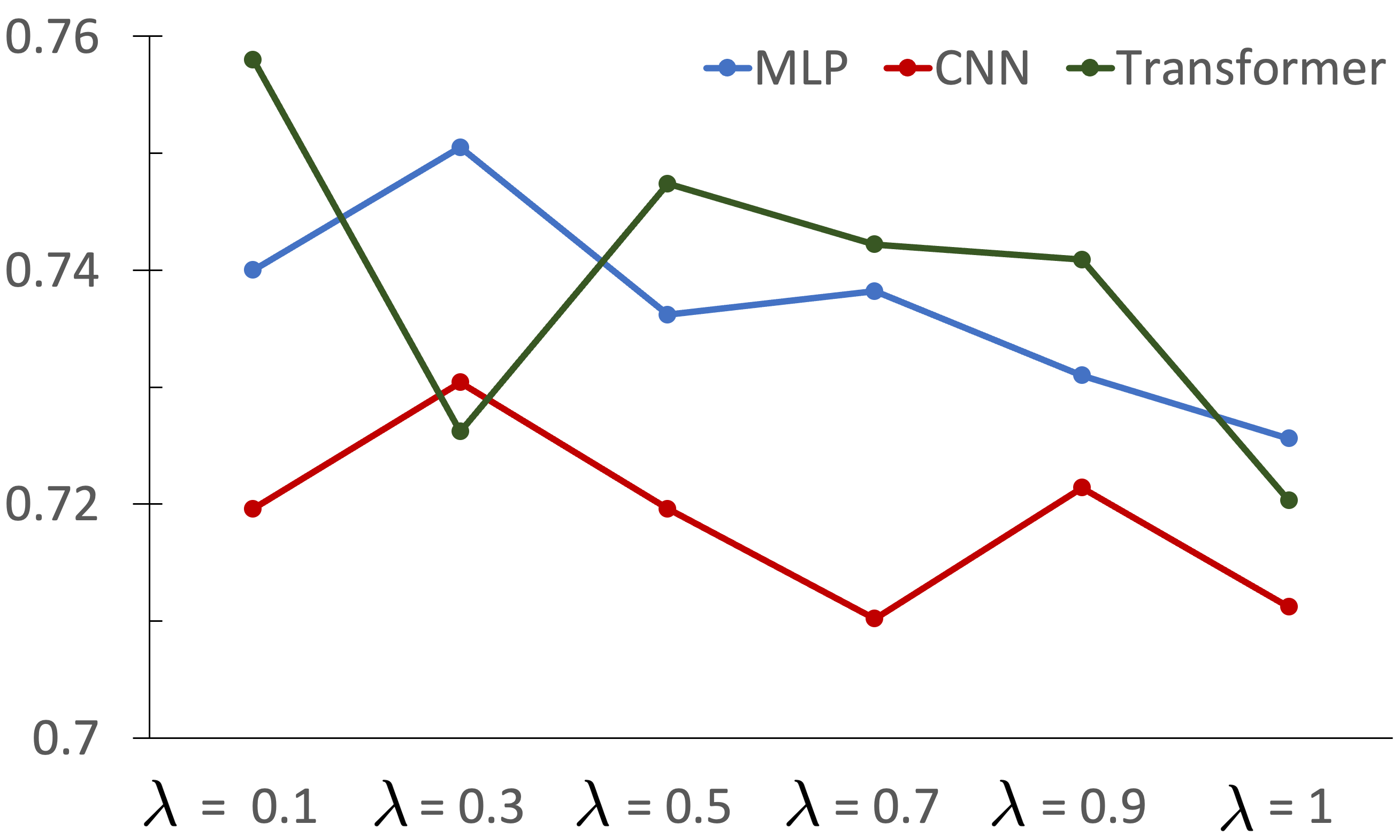}  
  \caption{GSE135304}
  \label{fig:ds_304}
\end{subfigure}
\begin{subfigure}[t]{.24\textwidth}
  \centering
  \includegraphics[width=1\linewidth]{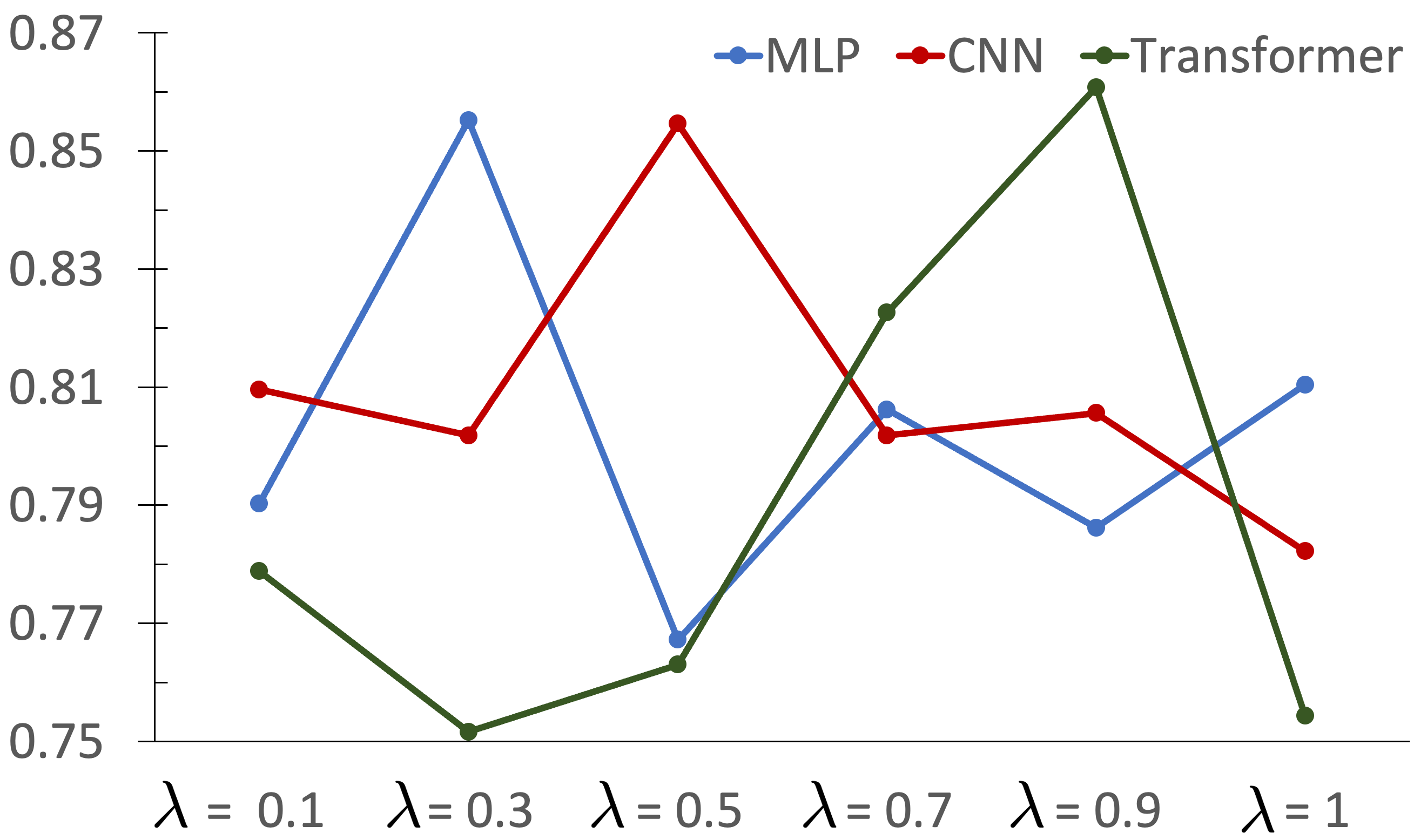}  
  \caption{GSE12771}
  \label{fig:ds_771}
\end{subfigure}
\begin{subfigure}[t]{.24\textwidth}
  \centering
  \includegraphics[width=1\linewidth]{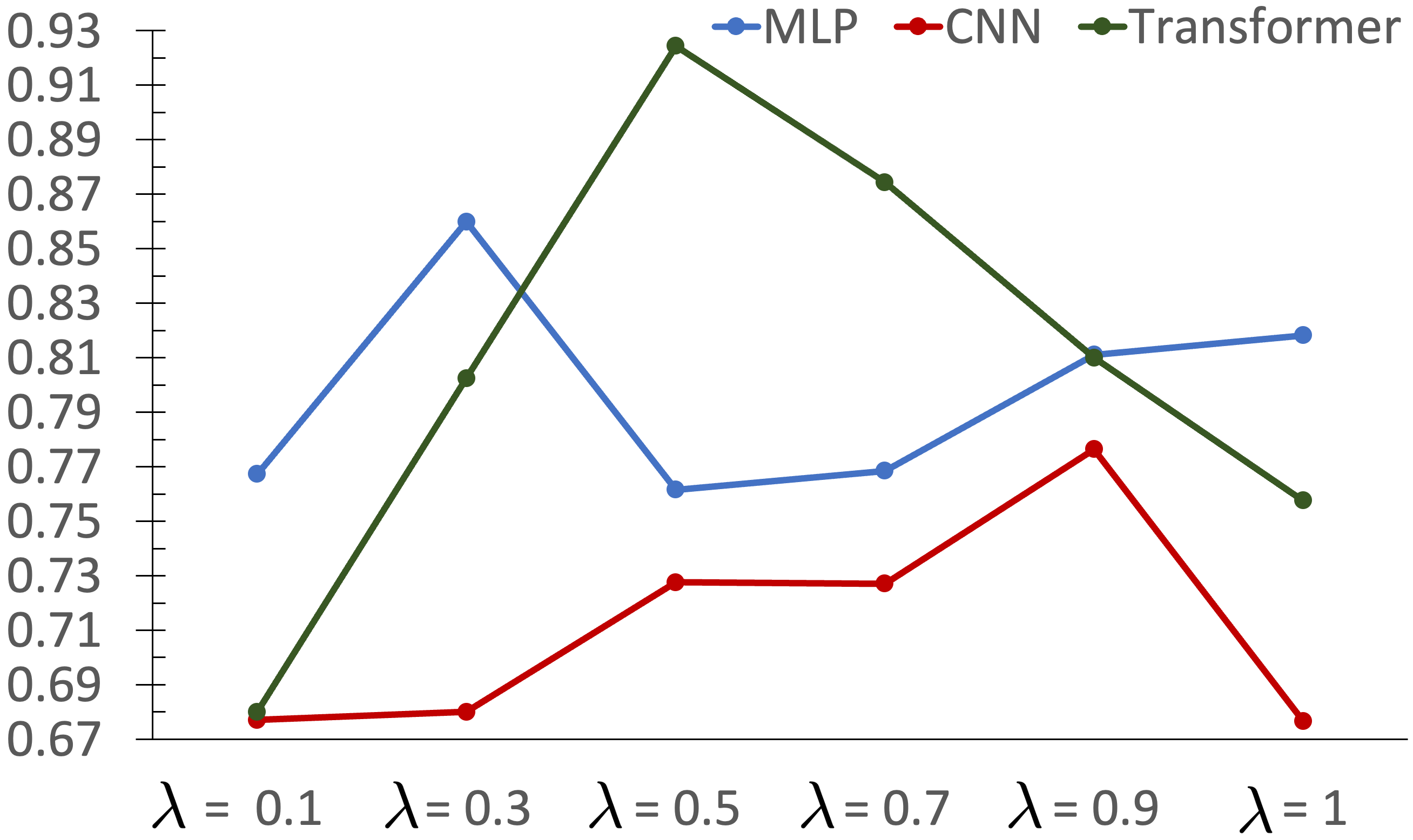}  
  \caption{GSE42830}
  \label{fig:ds_830}
\end{subfigure}
\vspace{0.2cm}
\caption{The effect of varying values of $\lambda$ in F1 score showing the influence of meta-learning. }
\label{fig:lambda}
\end{figure}

\subsection*{Results}
\begin{figure}[ht]
\centering
\begin{subfigure}[t]{.35\textwidth}
  \centering
  \includegraphics[width=\linewidth]{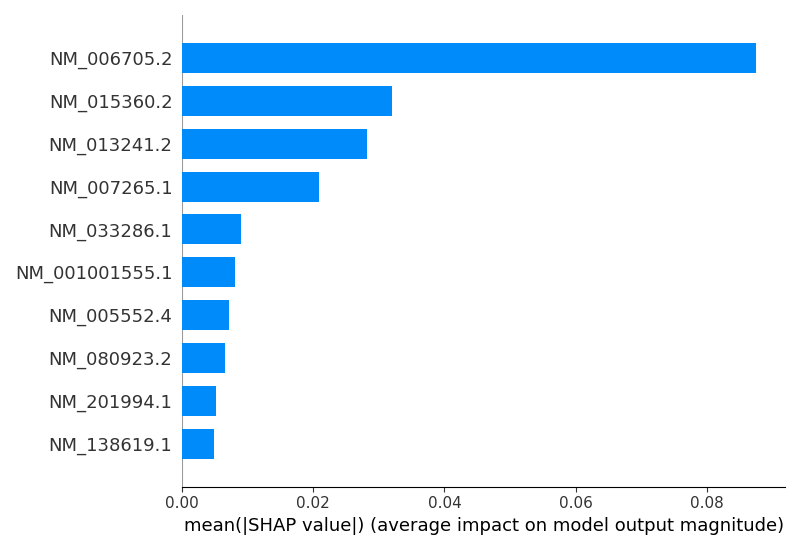}  
  \caption{Transformer + Meta-Learning}
  \label{fig:ds_255}
\end{subfigure}
\begin{subfigure}[t]{.35\textwidth}
  \centering
  \includegraphics[width=1\linewidth]{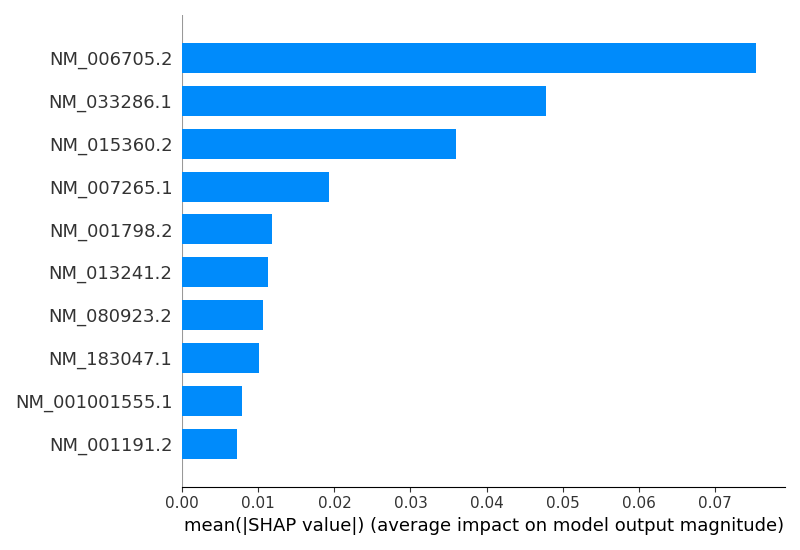}  
  \caption{Transformer w/o Meta-Learning}
  \label{fig:ds_304}
\end{subfigure}
\vspace{0.2cm}
\caption{Illustration of SHAP values for the test set of GSE135304 to find the most important gene sequences that effect the model decisions. The SHAP values with against without meta-learning for the Transformer are presented.}
\label{fig:shap}
\end{figure}

\textbf{RQ1. Comparison with Baseline Methods:} We evaluate the performance of our model across a variety of datasets to showcase the effectiveness of our proposed methodology. Notably, for methods without meta-learning, only the target datasets are utilized for optimization and sources are discarded. As shown in Tables \ref{table:255}, \ref{table:304}, \ref{table:771}, and \ref{table:830}, our meta-learning approach consistently surpasses traditional statistical methods, showing its efficacy in managing low-resource gene data. In all datasets, the Transformer's encoder outperformed both MLP and CNN architectures. 
Regarding statistical baselines, Random Forest and Support Vector Machines (SVM) demonstrate superior performance over decision trees and logistic regression because of their adaptability to high-dimensional feature spaces. Furthermore, our findings on datasets GSE12771 and GSE42830 show the ability of meta-learning to generalize effectively even when dealing with limited sample sizes (fewer than 100 instances). This is achieved by combining information from datasets with more samples. Thus, our results suggest that the proposed meta-learning approach is applicable to tackling the challenges of this domain.\\
\textbf{RQ2. Ablation Studies:} 
The meta-learning approach is employed in the optimization process of deep neural networks. To assess the effectiveness of this approach, we excluded meta-learning for MLP, CNN, and Transformer's encoders. As shown in Tables \ref{table:255}, \ref{table:304}, \ref{table:771}, and \ref{table:830}, meta-learning consistently enhances performance across all architectures. Particularly, The improvement is more pronounced on smaller datasets (Tables \ref{table:771} and \ref{table:830}) because of the utilization of information from larger datasets, which decreases the risk of overfitting issues and leads to a better generalization.\\
We also assess the level of meta-learning impact shown in Figure \ref{fig:lambda}. Across various datasets, we observe a performance peak for each architecture corresponding to a specific $\lambda$ value. This suggests a trade-off between leveraging solely the source data or incorporating the target data should be considered. A peak at associate $\lambda$ value near 0 indicates a stronger reliance on the source data. For instance, in the case of GSE135304 and  GSE13255, the peaks are at $\lambda = 0.1$ and $0.2$ showing that the distribution of the source data better suits the cancer detection task for this dataset. 
Additionally, we utilize SHAP values \cite{lundberg2017unified} to identify the important features contributing to the decision-making process. Figure \ref{fig:shap} illustrates the SHAP plot generated for the Transformers model applied to dataset GSE135304. Notably, the meta-learning approach changes the feature ranking and their respective impacts influenced by source datasets.\\
\textbf{RQ3. Transfer Learning:} Transfer learning \cite{transfer} has been introduced to leverage information from large datasets and fine-tune it to fit datasets with limited samples for better generalization. In transfer learning, a model is initially trained on source datasets and then fine-tuned on target datasets. We choose two datasets, GSE13255 and SE135304 for the transfer learning task as targets. As shown in Tables \ref{table:t255} and  \ref{table:t304}, meta-learning consistently outperforms transfer learning results in all of the datasets. This superiority is because of the fact that transfer learning heavily relies on the source task in the first stage and it separates the optimization for sources and the target. However, our approach includes the target task throughout the training process. Moreover, the results indicate that meta-learning exhibits greater generalization capabilities for cancer detection tasks based on gene expression levels compared to transfer learning.

\section*{Conclusion}
In our research, we present a meta-learning framework for genetic datasets with gene expression levels from DNA microarrays. This method improves neural network optimization by integrating similar domain datasets for enhanced generalization of smaller datasets and robust performance across diverse patient samples from four distinct datasets. Our approach outperforms statistical methods, deep learning techniques, and transfer learning across all datasets.

\section*{Acknowledgments}
This work was funded in part by the National Science Foundation (NSF) under award number DBI-1943291, and by the University of Kansas New Faculty Research Development (NFRD) Award.

\makeatletter
\renewcommand{\@biblabel}[1]{\hfill #1.}
\makeatother
\bibliographystyle{vancouver}
\bibliography{amia}  

\begin{thebibliography}{10}

\bibitem{cdc}
\text{Centers for Disease Control and Prevention (CDC)}. Lung Cancer
  Statistics; 2023.
\newblock Available from:
  \url{https://www.cdc.gov/lung-cancer/statistics/index.html}.

\bibitem{van2002gene}
Van't~Veer LJ, Dai H, Van De~Vijver MJ, He YD, Hart AA, Mao M, et~al.
\newblock Gene expression profiling predicts clinical outcome of breast cancer.
\newblock nature. 2002;415(6871):530-6.

\bibitem{burczynski2000toxicogenomics}
Burczynski ME, McMillian M, Ciervo J, Li L, Parker JB, Dunn RT, et~al.
\newblock Toxicogenomics-based discrimination of toxic mechanism in HepG2 human
  hepatoma cells.
\newblock Toxicological sciences. 2000;58(2):399-415.

\bibitem{staunton2001chemosensitivity}
Staunton JE, Slonim DK, Coller HA, Tamayo P, Angelo MJ, Park J, et~al.
\newblock Chemosensitivity prediction by transcriptional profiling.
\newblock Proceedings of the National Academy of Sciences.
  2001;98(19):10787-92.

\bibitem{ncbi}
\text{National Center for Biotechnology Information}. The HumanHT-12 v4
  Expression BeadChip; 2010.
\newblock Available from:
  \url{https://www.ncbi.nlm.nih.gov/geo/query/acc.cgi?acc=GPL10558}.

\bibitem{vadapalli2022artificial}
Vadapalli S, Abdelhalim H, Zeeshan S, Ahmed Z.
\newblock Artificial intelligence and machine learning approaches using gene
  expression and variant data for personalized medicine.
\newblock Briefings in bioinformatics. 2022;23(5):bbac191.

\bibitem{pividori2023projecting}
Pividori M, Lu S, Li B, Su C, Johnson ME, Wei WQ, et~al.
\newblock Projecting genetic associations through gene expression patterns
  highlights disease etiology and drug mechanisms.
\newblock Nature communications. 2023;14(1):5562.

\bibitem{data255}
Showe MK, Vachani A, Kossenkov AV, Yousef M, Nichols C, Nikonova EV, et~al.
\newblock Gene expression profiles in peripheral blood mononuclear cells can
  distinguish patients with non--small cell lung cancer from patients with
  nonmalignant lung disease.
\newblock Cancer research. 2009;69(24):9202-10.

\bibitem{data830}
Bloom CI, Graham CM, Berry MP, Rozakeas F, Redford PS, Wang Y, et~al.
\newblock Transcriptional blood signatures distinguish pulmonary tuberculosis,
  pulmonary sarcoidosis, pneumonias and lung cancers.
\newblock PloS one. 2013;8(8):e70630.

\bibitem{xie2016predictive}
Xie R, Quitadamo A, Cheng J, Shi X.
\newblock A predictive model of gene expression using a deep learning
  framework.
\newblock In: 2016 IEEE International Conference on Bioinformatics and
  Biomedicine (BIBM). IEEE; 2016. p. 676-81.

\bibitem{lopez2020transfer}
Lopez-Garcia G, Jerez JM, Franco L, Veredas FJ.
\newblock Transfer learning with convolutional neural networks for cancer
  survival prediction using gene-expression data.
\newblock PloS one. 2020;15(3):e0230536.

\bibitem{qiu2020meta}
Qiu YL, Zheng H, Devos A, Selby H, Gevaert O.
\newblock A meta-learning approach for genomic survival analysis.
\newblock Nature communications. 2020;11(1):6350.

\bibitem{finn2017model}
Finn C, Abbeel P, Levine S.
\newblock Model-agnostic meta-learning for fast adaptation of deep networks.
\newblock In: International conference on machine learning. PMLR; 2017. p.
  1126-35.

\bibitem{data304}
Kossenkov AV, Qureshi R, Dawany NB, Wickramasinghe J, Liu Q, Majumdar RS,
  et~al.
\newblock A gene expression classifier from whole blood distinguishes benign
  from malignant lung nodules detected by low-dose CT.
\newblock Cancer research. 2019;79(1):263-73.

\bibitem{data771}
Zander T, Hofmann A, Staratschek-Jox A, Classen S, Debey-Pascher S, Maisel D,
  et~al.
\newblock Blood-based gene expression signatures in non--small cell lung
  cancer.
\newblock Clinical Cancer Research. 2011;17(10):3360-7.

\bibitem{metalearning}
Nichol A, Achiam J, Schulman J.
\newblock On first-order meta-learning algorithms.
\newblock arXiv preprint arXiv:180302999. 2018.

\bibitem{biogrid}
Oughtred R, Rust J, Chang C, Breitkreutz BJ, Stark C, Willems A, et~al.
\newblock The BioGRID database: A comprehensive biomedical resource of curated
  protein, genetic, and chemical interactions.
\newblock Protein Science. 2021;30(1):187-200.

\bibitem{attentionall}
Vaswani A, Shazeer N, Parmar N, Uszkoreit J, Jones L, Gomez AN, et~al.
\newblock Attention is all you need.
\newblock Advances in neural information processing systems. 2017;30.

\bibitem{nayebi2023contrastive}
Nayebi~Kerdabadi M, Hadizadeh~Moghaddam A, Liu B, Liu M, Yao Z.
\newblock Contrastive learning of temporal distinctiveness for survival
  analysis in electronic health records.
\newblock In: Proceedings of the 32nd ACM International Conference on
  Information and Knowledge Management; 2023. p. 1897-906.

\bibitem{zhang2019metapred}
Zhang XS, Tang F, Dodge HH, Zhou J, Wang F.
\newblock Metapred: Meta-learning for clinical risk prediction with limited
  patient electronic health records.
\newblock In: Proceedings of the 25th ACM SIGKDD international conference on
  knowledge discovery \& data mining; 2019. p. 2487-95.

\bibitem{tan2022metacare++}
Tan Y, Yang C, Wei X, Chen C, Liu W, Li L, et~al.
\newblock Metacare++: Meta-learning with hierarchical subtyping for cold-start
  diagnosis prediction in healthcare data.
\newblock In: Proceedings of the 45th International ACM SIGIR Conference on
  Research and Development in Information Retrieval; 2022. p. 449-59.

\bibitem{lundberg2017unified}
Lundberg SM, Lee SI.
\newblock A unified approach to interpreting model predictions.
\newblock Advances in neural information processing systems. 2017;30.

\bibitem{transfer}
Weiss K, Khoshgoftaar TM, Wang D.
\newblock A survey of transfer learning.
\newblock Journal of Big data. 2016;3:1-40.

\end{thebibliography}

\end{document}